\documentclass[11pt, letterpaper]{article}
\usepackage[utf8]{inputenc}
\usepackage[margin=1.5cm]{geometry}
\usepackage{titlesec}
\usepackage{tabu}
\usepackage{enumitem}
\usepackage{amssymb}
\usepackage{xcolor}
\newlist{selectlist}{itemize}{2}
\setlist[selectlist]{label=$\square$,leftmargin=*,noitemsep,topsep=0pt}

\usepackage{graphicx}
\usepackage{caption}
\usepackage{bm}
\usepackage{cite}
\usepackage{adjustbox}

\usepackage{lmodern}

\usepackage{hyperref}
\hypersetup{
    colorlinks=true,
    linkcolor=blue,
    filecolor=magenta,      
    urlcolor=blue,
}
 
\titleformat{\section}[block]{\hspace{1em}\bfseries}{\thesection.}{0.5em}{} 
\titleformat{\subsection}[block]{\hspace{1em}}{\thesubsection}{0.5em}{}

\usepackage{indentfirst}
\begin{document}

\noindent
\textbf{GP CC-OPF: Gaussian Process based optimization tool for Chance-Constrained Optimal Power Flow}
\vskip0.5cm
\noindent

\noindent
\textbf{Mile Mitrovic}, Center for Energy Science and Technology, Skolkovo Institute of Science and Technology, Moscow, Russia, mile.mitrovic@skoltech.ru\\
\textbf{Ognjen Kundacina}, The Institute for Artificial Intelligence Research and Development of Serbia, Novi Sad, Serbia, ognjen.kundacina@ivi.ac.rs\\
\textbf{Aleksandr Lukashevich}, Center for Energy Science and Technology, Skolkovo Institute of Science and Technology, Moscow, Russia, aleksandr.lukashevich@skoltech.ru\\
\textbf{Petr Vorobev}, Center for Energy Science and Technology, Skolkovo Institute of Science and Technology, Moscow, Russia, p.vorobev@skoltech.ru\\
\textbf{Vladimir Terzija}, Center for Energy Science and Technology, Skolkovo Institute of Science and Technology, Moscow, Russia, v.terzija@skoltech.ru\\
\textbf{Yury Maximov}, Theoretical Division, Los Alamos National Laboratory, Los Alamos NM, USA, yury@lanl.gov\\
\textbf{Deepjyoti Deka}, Theoretical Division, Los Alamos National Laboratory, Los Alamos NM, USA, deepjoyti@lanl.gov\\

\noindent
\textbf{Abstract}\\
\textit{The Gaussian Process (GP) based Chance-Constrained Optimal Power Flow (CC-OPF) is an open-source Python code developed for solving economic dispatch (ED) problem in modern power grids. In recent years, integrating a significant amount of renewables into a power grid causes high fluctuations and thus brings a lot of uncertainty to power grid operations. This fact makes the conventional model-based CC-OPF problem non-convex and computationally complex to solve. The developed tool presents a novel data-driven approach based on the GP regression model for solving the CC-OPF problem with a trade-off between complexity and accuracy. The proposed approach and developed software can help system operators to effectively perform ED optimization in the presence of large uncertainties in the power grid.}
\vskip0.5cm

\vskip0.5cm
\newpage
\noindent
\textbf{Code metadata}\\

\noindent
\begin{tabular}{|l|p{6.5cm}|p{9.5cm}|}
\hline
\textbf{Nr.} & \textbf{Code metadata description} & \textbf{Please fill in this column} \\
\hline
C1 & Current code version & $v1.0.0$ \\
\hline
C2 & Permanent link to code/repository used for this code version & https://github.com/mile888/hybrid\_gp \\
\hline
C3  & Permanent link to Reproducible Capsule & https://codeocean.com/capsule/9532852/tree/v1  \\ 
\hline
C4 & Legal Code License   & MIT \\
\hline
C5 & Code versioning system used & none \\
\hline
C6 & Software code languages, tools, and services used & Python\\
\hline
C7 & Compilation requirements, operating environments \& dependencies &  NumPy, Pandas, SciPy, CasADi\\
\hline
C8 & If available Link to developer documentation/manual & \\
\hline
C9 & Support email for questions & mile.mitrovic@skoltech.ru\\
\hline
\end{tabular}\\
\vskip0.5cm
\noindent

\section{Introduction}
\label{sec: intro}
The optimal power flow (OPF) plays an important role for the secure and economic operation of the power grids. As an optimization tool, the OPF is typically used to solve the Economic dispatch (ED) problem by finding the optimal output of the controllable generators with the lowest possible cost that meets the load and physical constraints of the grid. However, the OPF is a complex non-linear problem with many constraints that can be hard to solve. In addition, the rapid integration of renewable energy resources (RES) with intermittent outputs propagates uncertainty through the grid and thus leads to a higher degree of complexity in power grid operations. To take into account the impacts of uncertainty within the OPF, the researchers have recently proposed several stochastic approaches such as robust 
optimization \cite{fang2019distributionally}, probabilistic OPF \cite{cao2017probabilistic},
and Chance-Constrained (CC) OPF \cite{lubin2019chance, viafora2020chance}. Robust optimization often leads to conservative solutions, while probabilistic OPF is difficult to implement in practice. The CC-OPF implies satisfying probability constraints with a given acceptable violation probability, balancing operating costs and security in the power grid in that way. 

In this paper, we put focus on the practical CC-OPF approach. However, model-based CC-OPF is notoriously hard to solve for the full non-linear alternating current (AC) power flow model. To overcome this problem, the novel data-driven approach based on Gaussian Process (GP) regression model is proposed \cite{mile1, mile2}. The proposed approach uses a hybrid GP model by combining a linear model, based on the direct current (DC) power flow, with the data-driven estimation of the residuals between DC and AC power flow models. Thus, we provide a more robust and accurate GP CC-OPF. To provide a general and scalable approach, applicable to a larger power grid, we have added a sparse GP option to select the most important data in the feature space for hybrid GP estimation. Accordingly, the computational complexity is significantly reduced without compromising the accuracy of the solution.

Finally, we provide Python code to simulate and run the proposed GP CC-OPF approach. The goal of the software is to provide a tool for performing stochastic OPF under uncertain generation resources that can be used both by the industry and academia.

\section{Description of GP CC-OPF}
The GP CC-OPF approach has been implemented in Python. The general scheme of the GP CC-OPF approach is shown in Fig. \ref{fig: CodeWorkflow}. Code details are presented below.  

\begin{figure*}
    \centering
    \captionsetup{justification=centering,font=footnotesize}
    \includegraphics[width=0.90\textwidth]{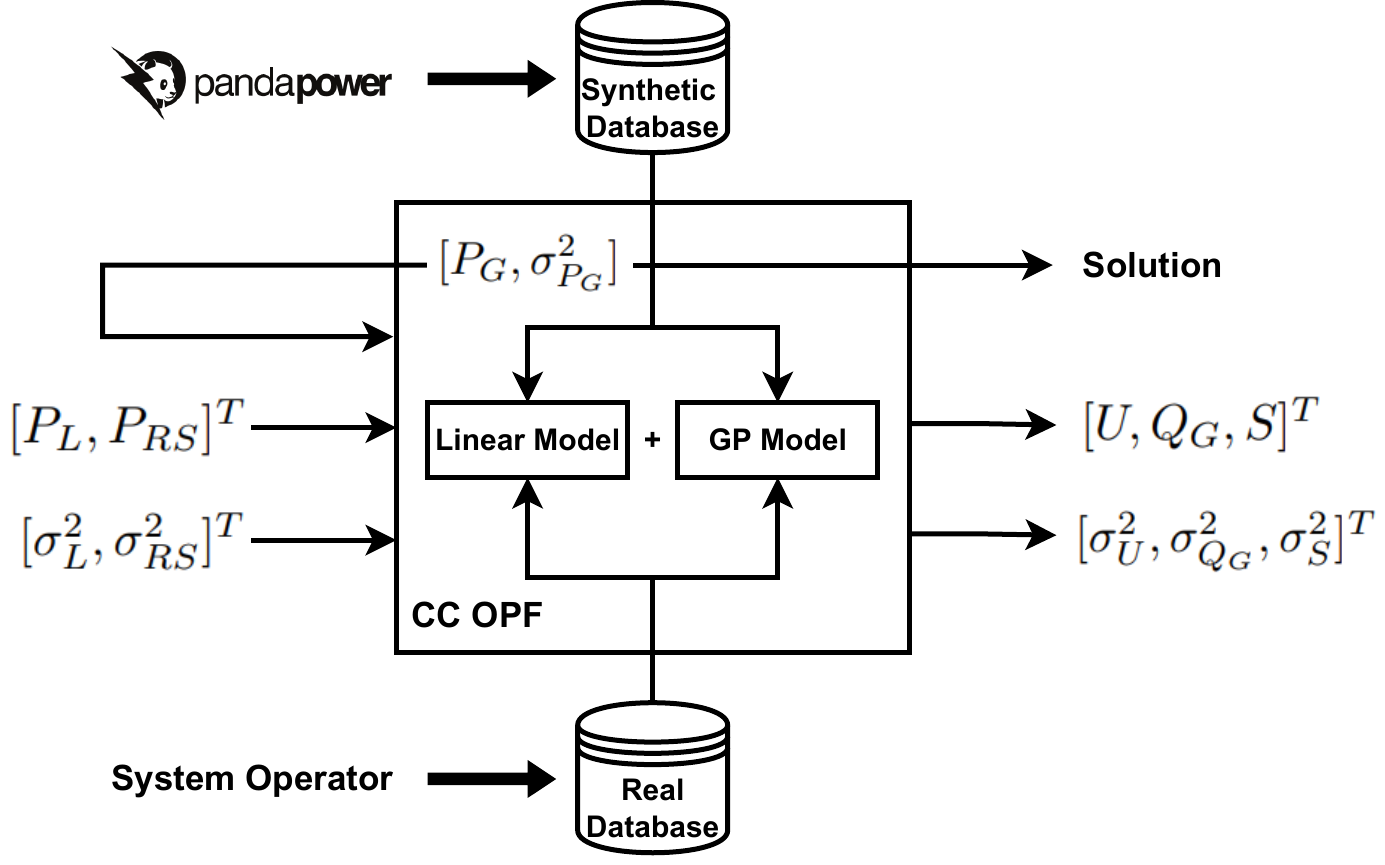}
    \caption{General scheme of GP CC-OPF approach.}
    \label{fig: CodeWorkflow}
\end{figure*}

\subsection{Data preparation}
There are two ways to provide labeled data for training the Linear and GP regression models. The first option is using synthetic data by modeling the power grid. This tool uses \texttt{pandapower} \cite{thurner2018pandapower} Python library for simulating IEEE9 and IEEE39-bus power grids. The second option is to use data from the real power grid provided by the system operator.

In both cases, the GP model is learned on standardized data. The data sent to the GP model consists of the residuals between datasets obtained from the AC and DC power flow models. The linear model illustrates the ordinary least squares Linear Regression model provided in \texttt{scikit-learn} library \cite{scikit-learn}, and it uses data generated from a linear DC power flow model.  

The SLSQP \cite{kraft1988software} non-linear solver is used to minimize the objective function and find the optimal hyperparameters of the GP model. In the case of sparse GP approximation, variational learning is applied to find inducing points and better approximate marginal likelihood \cite{titsias2009variational}. 

\subsection{GP CC-OPF setup}
The GP CC-OPF approach presents the conventional chance-constrained optimization problem with an integrated data-driven machine learning model, as shown in Fig.~\ref{fig: CodeWorkflow}. The data-driven model replaces the standard non-linear AC power flow equations. To make the task more accurate, the data-driven model consists of a combination of two models: a linear model and an additive GP model. Considering this, hybrid GP model, we make the approximated model more robust and in the worst case, the hybrid GP model will be equal to the standard DC convex power flow model. Moreover, GP can fit any smooth non-linear function very well and thus avoid some local minima. This fact can simplify the solution of the CC-OPF problem by ensuring that the engineering constraints are reliably feasible for any uncertainty realization.

The advantage of the proposed hybrid GP CC-OPF approach is that there is no need to know the grid configuration and parameters. Thus, problems with uncertain or unreliable
power system parameters encountered by conventional model-based algorithms are avoided. This makes the proposed data-driven approach more robust. As input data, it is enough for the user to provide the active power of the load and RES, as well as their standard deviations as uncertainty factors. The input uncertainties are propagated to output variables and their impact is modeled using the first-order Taylor approximation (TA1) by linearizing the data-driven function at the decision point. The active power and uncertainty factor of the controllable generators are optimized according to the cost function and participation factor, which are also considered as inputs in the data-driven model.

The algorithm is implemented in the non-linear optimization framework CasADi \cite{andersson2019casadi} using the available IPOPT solver \cite{wachter2006implementation} to solve the optimization problem.

\subsection{Outputs of the tool}
Based on the results of the GP CC-OPF and the developed tool, the user obtains the optimal solution of controllable generators with the optimal cost. Moreover, the tool provides output variables and their propagated standard deviations, such as voltage magnitude at non-controllable buses, reactive power at controllable buses and apparent power flow in the lines of the grid. Accordingly, the user can monitor whether the solution is feasible and accurate.

\subsection{Approach validation}
\begin{figure}[ht]  
 \begin{minipage}{0.3\textwidth}
   \includegraphics[width=\textwidth]{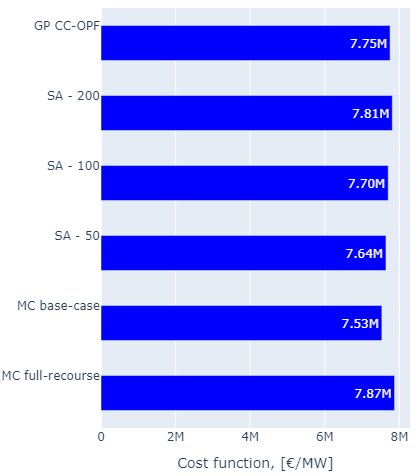}
  \end{minipage}
  \hfill
  \begin{minipage}{0.3\textwidth}
   \includegraphics[width=\textwidth]{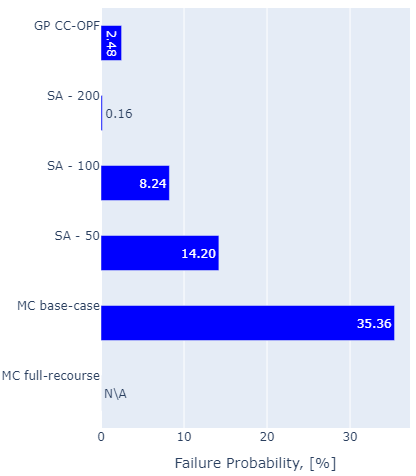}
  \end{minipage}
  \hfill
  \begin{minipage}{0.3\textwidth}
   \includegraphics[width=\textwidth]{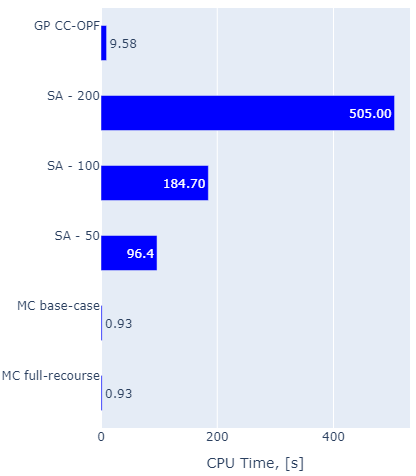}
  \end{minipage}
  \caption{Cost function values, probability of a failure and CPU time for IEEE9 (red) and IEEE39 (blue) systems.}
  \label{fig: tables}
\end{figure}

For performance evaluation, we compared the proposed data-driven approach with two alternative sample-based reformulations of the CC-OPF based on the scenario approach (SA) with a different number of scenarios and the Monte Carlo (MC) simulation (full-resources and base-case). In Fig.~\ref{fig: tables}, we present the values of the cost function, CPU time, and probability of failure for both IEEE9 and IEEE39-bus systems. The simulations are performed with the acceptable violation probabilities of all constraints set to $2.5\%$. The numerical evaluations from Fig.~\ref{fig: tables} show that the proposed hybrid GP CC-OPF provides a trade-off between cost and computational effort without sacrificing the accuracy of the given violation limits.

\section{Software impacts}
At the moment, there is no developed software in the industry that solves an uncertainty-aware OPF. System operators still use deterministic OPF solutions without considering generation uncertainties. As for the scientific community, a significant amount of work has been done in the field of stochastic OPF, but the algorithms proposed in the literature are typically not available to the public. Accordingly,  researchers face difficulties in comparing the performance of newly proposed methods with the existing work. Moreover, it requires additional effort to re-implement the existing algorithms, as we have done in our recent work \cite{mile1, mile2}. Because of these facts, the proposed open-source code can have a significant impact on both industry and academia.

In future versions, we plan to provide this tool with a friendly user interface that is easy to use. In the current version, the proposed tool works with IEEE9 and 39-bus grids. Our next task will be to make the existing tool universal for working with any scale of the system. 

We firmly believe that the proposed data-driven approach in the form of a public code will have a positive impact on future research in the field of stochastic OPF and motivate engineers to apply it in the industry.
\\

\begin{figure}[ht]  
 \begin{minipage}{0.3\textwidth}
   \includegraphics[width=\textwidth]{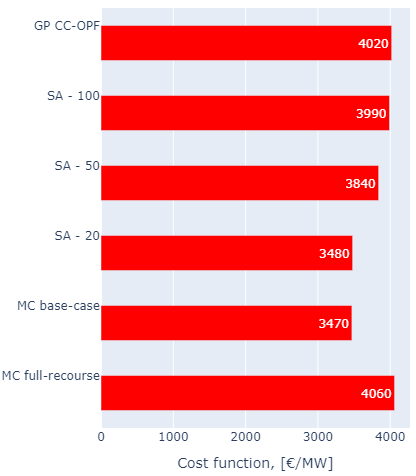}
  \end{minipage}
  \hfill
  \begin{minipage}{0.3\textwidth}
   \includegraphics[width=\textwidth]{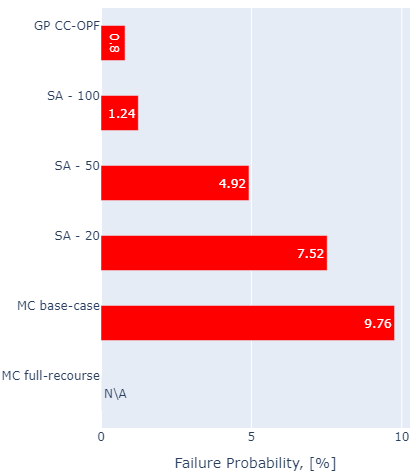}
  \end{minipage}
    \hfill
  \begin{minipage}{0.3\textwidth}
   \includegraphics[width=\textwidth]{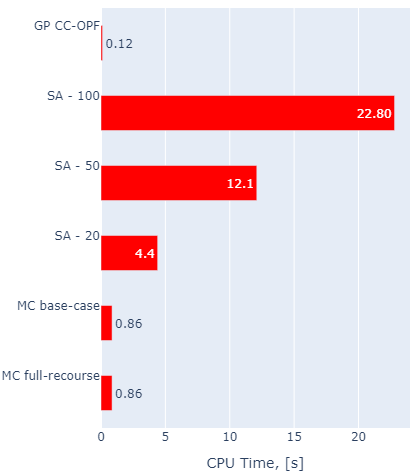}
  \end{minipage}
\end{figure}


\noindent
\bibliographystyle{IEEEtran}
\bibliography{main.bib}



\end{document}